\documentclass[sigconf,screen,authorversion,nonacm]{acmart}

\usepackage{graphicx}
\usepackage{booktabs}
\usepackage{multirow}
\graphicspath{{Image/}}
\usepackage{enumitem}
\usepackage[table]{xcolor}
\usepackage[most]{tcolorbox}
\tcbuselibrary{listings,skins,breakable}

\setcopyright{acmlicensed}
\copyrightyear{2026}
\acmYear{2026}
\acmDOI{10.1145/XXXXXX.XXXXXX} 
\acmISBN{978-1-4503-XXXX-X/26/06}
\acmSubmissionID{}    
\usepackage{listings}  

\definecolor{promptbg}{RGB}{243,248,252}
\definecolor{promptframe}{RGB}{119,136,153}
\definecolor{prompttitle}{RGB}{42,88,132}

\lstdefinestyle{promptstyle}{
  basicstyle=\ttfamily\bfseries\small,
  breaklines=true,
  columns=fullflexible,
  keepspaces=true,
  showstringspaces=false,
  frame=none,
  xleftmargin=0pt,
  xrightmargin=0pt,
  aboveskip=0pt,
  belowskip=0pt,
}

\newtcblisting{promptpanel}[2][]{%
  enhanced,
  colback=promptbg,
  colframe=promptframe,
  boxrule=0.6pt,
  arc=1mm,
  left=3mm,
  right=3mm,
  top=4mm,
  bottom=3mm,
  listing only,
  listing engine=listings,
  listing options={style=promptstyle,#1},
  boxed title style={
    colback=prompttitle,
    colframe=prompttitle,
    boxrule=0pt,
    arc=1mm,
    left=8pt,
    right=8pt,
    top=4pt,
    bottom=4pt,
  },
  fonttitle=\bfseries\small,
  coltitle=white,
  attach boxed title to top center={yshift=-2.5mm},
  title={#2},
}

\renewcommand\footnotetextcopyrightpermission[1]{}

\vspace{-1em}

\title{GRASP: Grounded CoT Reasoning with Dual-Stage Optimization for Multimodal Sarcasm Target Identification}

\author{Faxian Wan}
\affiliation{%
  \institution{School of Computer Science and Engineering, Northeastern University}
  \city{Shenyang}
  \postcode{110819}
  \country{China}
}
\email{wanfaxian@mails.neu.edu.cn}

\author{Xiaocui Yang}\authornote{Corresponding author.}
\affiliation{%
  \institution{School of Computer Science and Engineering, Northeastern University}
  \city{Shenyang}
  \postcode{110819}
  \country{China}
}
\email{yangxiaocui@cse.neu.edu.cn}

\author{Yifan Cao}
\affiliation{%
  \institution{School of Computer Science and Engineering, Northeastern University}
  \city{Shenyang}
  \postcode{110819}
  \country{China}
}
\email{caoyifan@mails.neu.edu.cn}

\author{Shi Feng}
\affiliation{%
  \institution{School of Computer Science and Engineering, Northeastern University}
  \city{Shenyang}
  \postcode{110819}
  \country{China}
}
\email{fengshi@cse.neu.edu.cn}

\author{Daling Wang}
\affiliation{%
  \institution{School of Computer Science and Engineering, Northeastern University}
  \city{Shenyang}
  \postcode{110819}
  \country{China}
}
\email{wangdaling@cse.neu.edu.cn}

\author{Yifei Zhang}
\affiliation{%
  \institution{School of Computer Science and Engineering, Northeastern University}
  \city{Shenyang}
  \postcode{110819}
  \country{China}
}
\email{zhangyifei@cse.neu.edu.cn}

\begin{document}

\begin{abstract}
Moving beyond the traditional binary classification paradigm of Multimodal Sarcasm Detection, Multimodal Sarcasm Target Identification (MSTI) presents a significantly more formidable challenge. It necessitates the precise localization of fine-grained targets, such as specific textual phrases and visual regions, to achieve a nuanced understanding of cross-modal sarcasm. Existing approaches predominantly rely on implicit cross-modal alignment, suffering from a lack of interpretability and yielding suboptimal performance in fine-grained target localization. To address these limitations, we propose \textbf{GRASP}, \textbf{G}rounded Chain-of-Thought \textbf{R}e\textbf{A}soning with Dual-Stage Optimization for Multimodal \textbf{S}arcasm \textbf{P}rediction and Target Identification, a novel framework that pioneers the integration of visual grounding and explicit Chain-of-Thought (CoT) reasoning to move beyond black-box MSTI. Specifically, we curate \textbf{MSTI-MAX}, a refined dataset that mitigates class imbalance and enriches multimodal sarcasm cues. To endow GRASP with human-like intent decomposition, we introduce \textbf{Grounded CoT reasoning}, which explicitly anchors sarcasm-related visual regions within the text reasoning trajectory. It prompts the model to explicitly articulate the rationale behind the sarcastic intent, grounded in multimodal perception and reasoning, prior to predicting the final classification labels and sarcasm targets. Furthermore, the dual-stage outcome-supervised joint optimization strategy combining Supervised Fine-Tuning with coordinate-aware weighted loss, and Fine-Grained Target Policy Optimization is employed to solidify the robust reasoning capabilities of the model. Extensive experiments demonstrate that GRASP outperforms existing baselines in fine-grained sarcasm target identification across different modalities. Notably, beyond traditional metrics that merely assess predictive correctness, we introduce an LLM-as-a-Judge evaluation to quantitatively measure the quality of the internal reasoning chains. Our work paves the way for explicit grounding reasoning in complex, fine-grained multimodal tasks. 
Our built dataset and source code will be released on GitHub.
\end{abstract}

\maketitle
\vspace{-0.5em}
\section{Introduction}

\begin{figure}[t]
    \centering
    \Description{An example of multimodal sarcasm. The text expresses surprise with heavy irony, while the image shows a dangerously low-flying aircraft over highway traffic, creating a strong text-image semantic contradiction.}   
    \includegraphics[width=0.9\linewidth]{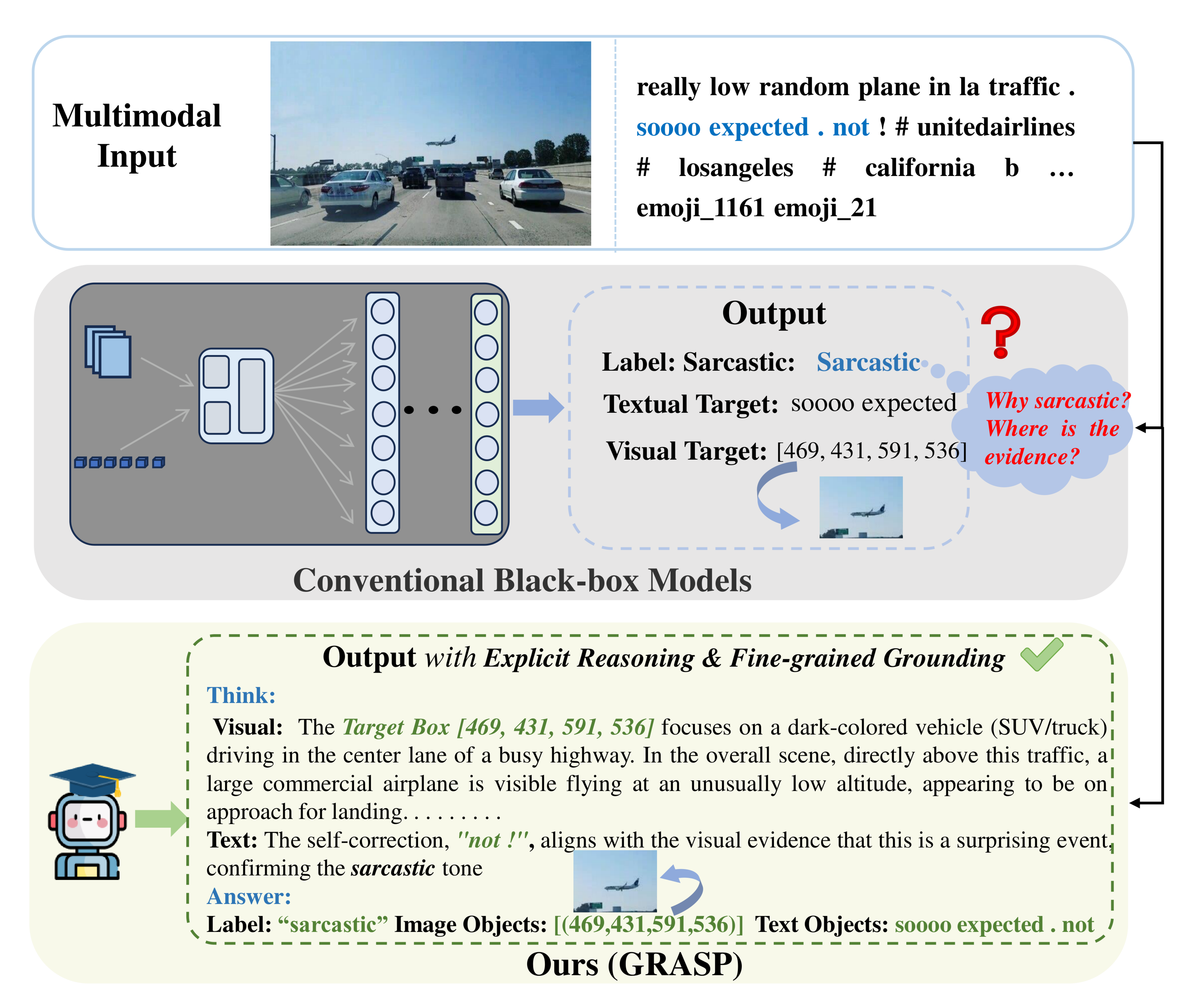}
    \vspace{-1em}
    \caption{An illustrative example of MSTI. The sarcasm arises from the contradiction between the specific textual span “soooo expected. not” and the localized visual region (depicting a perilously low-flying airplane), which collectively conveys the sarcastic intent. Our GRASP framework not only outputs fine-grained multimodal targets directly but also yields explicit grounding chain-of-thought reasoning. }
    \vspace{-1.5em}
    \label{Fig:1_intro}
\end{figure}
Despite progress in Multimodal Sarcasm Detection (MSD), existing studies primarily formulate it as a macroscopic binary classification task, predicting merely whether a given image-text pair is sarcastic ~\cite{schmsd2016, castromsdvido2019, cai2019multi, 2025context-msd, zhou2026MSD}. However, this coarse-grained paradigm falls fundamentally short in scenarios that demand deep interpretability and fine-grained localization. To bridge this gap, the Multimodal Sarcasm Target Identification (MSTI) ~\cite{wang2022multimodal} task is introduced. MSTI requires models to explicitly ground the specific targets that trigger sarcasm, such as precise textual spans and their corresponding visual bounding boxes. As illustrated in Figure \ref{Fig:1_intro}, beyond merely detecting the presence of sarcasm, a model must accurately localize the fine-grained triggers, such as the sarcastic textual cue, i.e., \textit{“soooo expected. not”}, and the specific image region containing \textit{the abnormally low-flying plane}. The paradigm shift from holistic detection to fine-grained identification poses formidable challenges to multimodal understanding, reasoning, and alignment.

Existing research on MSD and MSTI remains constrained by critical challenges.
\textbf{First, the lack of explicit reasoning and interpretability due to black-box perception paradigms.} Although recent models for MSTI ~\cite{chen2024cofipara, lv2025mstiplus} attempt to incorporate fine-grained visual bounding boxes to localize sarcastic targets, they predominantly rely on the task-specific trained architecture or cumbersome multi-stage pipeline. These approaches typically produce classification labels and spatial coordinates directly via implicit feature fusion and large-scale pre-training, failing to capture the logical reasoning trajectory from multimodal perception and reasoning to the final detection, as shown in Figure \ref{Fig:1_intro}. Consequently, even when predictions are accurate, it remains ambiguous whether the model has genuinely comprehended the underlying cross-modal incongruity or merely overfitted to spurious correlations in the dataset.
\textbf{Second, the inadequacy of general-purpose models in tackling the complex, fine-grained grounding task.} While recent Multimodal Large Language Models (MLLMs), e.g., GPT-4o ~\cite{openai2023gpt4v}, Qwen-VL ~\cite{bai2025qwen25vl}, exhibit remarkable reasoning capabilities, empirical evaluations ~\cite{wang2025can, chen2024cofipara} reveal their sub-optimal performance on the fine-grained sarcasm detection task. The phenomenon stems from the fact that sarcasm heavily relies on implicit cultural nuances and pragmatic reversals, which evade general pre-training. Furthermore, most existing research relegates MLLMs to the role of an auxiliary knowledge generator ~\cite{wei2025deepmsd, chen2024cofipara, zhao2025eilmob}. During the reasoning process, they typically compress complex multimodal signals into discrete textual representations, such as image captions or text tags, thereby degrading fine-grained visual details. Consequently, they fail to synergistically integrate the cognitive rationale (\textit{why it is sarcastic}) with fine-grained visual grounding (\textit{where the sarcasm lies}) within a unified, end-to-end generative framework.

To address these challenges, a high-quality fine-grained sarcasm target identification dataset is needed. However, we reveal that the fine-grained sarcasm dataset, MSTI~2.0 ~\cite{chen2024cofipara}, is hindered by data imbalance and the lack of annotations for salient visual sarcasm targets. To bridge this gap, we reconstruct and refine MSTI~2.0 to build the new dataset, \textbf{MSTI-MAX},  with grounded multimodal reasoning.
MSTI-MAX not only provides bounding boxes for visual targets and key textual spans for fine-grained sarcasm target identification but also achieves balanced data distribution, laying the robust foundation.
To address the aforementioned challenges, we propose \textbf{GRASP}, Grounded Chain-of-Thought Reasoning with Dual-Stage Optimization for Multimodal Sarcasm Prediction and Target Identification. Specifically, GRASP enforces structured CoT instructions, where within a dedicated \texttt{Think} module, the model first extracts salient visual anchors via bounding boxes and core text spans for fine-grained perception, and then explicitly aligns them to mitigate cross-modal incongruities. By directly referencing visual regions during reasoning, rather than compressing them into textual representations, GRASP effectively achieves ``Thinking with Images'' ~\cite{su2025thinking,li2025thinkimage}. It preserves fine-grained spatial details and fosters cooperation between visual grounding and abstract cognitive reasoning. Following this explicit reasoning process, the model yields the classification label, textual targets, and precise bounding box coordinates within a final \texttt{Answer} module.
To endow GRASP with robust reasoning capabilities, we devise the novel \textbf{dual-stage outcome-supervised joint optimization} strategy. In the first stage, Supervised Fine-Tuning (SFT) is employed to distill high-quality CoT trajectories from a powerful MLLM. Concurrently, we introduce a \textbf{coordinate-aware weighted loss} to amplify gradients on bounding box tokens, thereby enhancing visual grounding performance. In the second stage, we propose \textbf{Fine-Grained Target Policy Optimization (FTPO)} to further refine the reasoning chains. FTPO leverages multi-dimensional rewards that encompass formatting, classification accuracy, visual grounding IoU, and textual target matching. Our optimization empowers the model to autonomously self-improve its cross-modal incongruity perception and target localization precision.

Furthermore, traditional outcome-oriented metrics, e.g., Accuracy, F1-score, fail to capture the nuanced cognitive processes underlying sarcasm comprehension. We introduce the LLM-as-a-Judge~\cite{zheng2023judging} to evaluate the quality of reasoning for MSTI.
The advanced model is leveraged to quantify intermediate reasoning trajectories across three dimensions, including \textit{Visual Perception}, \textit{Incongruity Reasoning}, and \textit{Logical Consistency}. Extensive experiments demonstrate that our framework achieves state-of-the-art (SOTA) performance on MSTI and exhibits competitive performance in sarcasm detection. 
Our contributions are summarized as follows:
\begin{enumerate}[leftmargin=13pt]
    \item We introduce \textbf{MSTI-MAX}, a reconstructed and balanced fine-grained dataset tailored for Multimodal Sarcasm Target Identification (MSTI). By incorporating detailed reasoning rationales, it establishes a valuable benchmark for the community.
    \item We propose \textbf{GRASP}, a novel end-to-end framework driven by visual Grounded Chain-of-Thought (CoT) reasoning. It seamlessly unifies explicit reasoning, fine-grained visual localization, and cross-modal incongruity resolution.
    \item To endow GRASP with robust reasoning capabilities, we devise a novel \textbf{dual-stage outcome-supervised optimization strategy}, which integrates SFT with a coordinate-aware weighted loss, followed by FTPO with multiple fine-grained rewards.
    \item 
    We conduct comprehensive experiments and pioneer the LLM-as-a-Judge evaluation to quantitatively evaluate the intermediate reasoning trajectories.
\end{enumerate}
\vspace{-1em}
\section{Related Works}
\subsection{MLLMs for Multimodal Sarcasm}
The emergence of Multimodal Large Language Models (MLLMs), such as InternVL~\cite{chen2024internvl} and Qwen-VL~\cite{bai2025qwen25vl}, has opened new avenues for multimodal sarcasm. Several recent works have explored the potential of MLLMs in zero-shot or few-shot sarcasm tasks~\cite{tang2024leveraging}. For instance, Chen et al.~\cite{chen2025seeing} evaluate multimodal sarcasm perception across 12 MLLMs, highlighting the profound challenges posed by sarcastic ambiguity. Similarly, Wang et al.~\cite{wang2025can} evaluate MLLMs on sarcasm detection and explanation, identifying core limitations such as insufficient visual comprehension and a lack of conceptual knowledge. 
It is primarily because sarcasm necessitates a fine-grained understanding of cultural contexts, emotional inconsistencies, and implicit communicative intentions, nuances that are difficult to capture through general pre-training alone~\cite{llm-msd}.
To bridge this gap, fine-tuning approaches have shown greater potential. For example, Emotion-EilMoB~\cite{zhao2025eilmob} integrates MLLM-generated textual knowledge into the model to enhance comprehension. Furthermore, CoT reasoning~\cite{wang2023selfconsistency} has been introduced to facilitate structured textual deductions. Recently, Su et al.~\cite{su2025thinking} proposed the ``Thinking with Images'' paradigm, which enables models to utilize visual information as intermediate reasoning steps, thereby transforming vision from a passive input into a dynamic, actionable cognitive workspace~\cite{huang2025vision, bai2025univg}. 
Departing from previous works that evaluate them in constrained zero-shot settings~\cite{wang2025ksdgcn, jana2025think} or merely utilize MLLMs as external knowledge generators, we propose \textbf{GRASP}. It performs end-to-end training on MLLMs using our curated \textbf{MSTI-MAX} dataset, tailored for MSD and MSTI. 

\vspace{-1em}
\subsection{Multimodal Sarcasm Target Identification}

Multimodal Sarcasm Detection (MSD)~\cite{schmsd2016} aims to determine whether a given image-text pair conveys sarcastic intent. Pioneering this field, Cai et al.~\cite{cai2019multi} introduce MMSD, the large-scale multimodal sarcasm dataset sourced from Twitter. Subsequently, Qin et al.~\cite{qin2023mmsd2} refine this benchmark into MMSD~2.0 by eliminating spurious cues and correcting erroneous labels. Despite these advancements, MSD remains confined to coarse-grained binary classification. It fails to localize the specific textual or visual elements that trigger the sarcasm, a capability indispensable for achieving a nuanced and interpretable understanding of sarcastic expressions.
To address this limitation, Multimodal Sarcasm Target Identification (MSTI)~\cite{wang2022multimodal} is introduced to pinpoint the exact sarcastic entities across modalities. The effort yields the MSTI dataset, which provides fine-grained annotations including textual target spans and visual bounding boxes. Building upon this foundation, Chen et al.~\cite{chen2024cofipara} develop  MSTI~2.0 and propose CofiPara, employing a coarse-to-fine training paradigm guided by contrastive rationales generated by MLLMs, like Qwen-VL~\cite{bai2025qwen25vl}. More recently, Lv et al.~\cite{lv2025mstiplus} construct the MSTI-Plus benchmark by incorporating fine-grained non-sarcastic aspect annotations and propose the Semantic-aware Sarcasm Target Identification (SaSTI) module. 

However, current MSTI research still faces two critical bottlenecks. First, MSTI~2.0 suffers from suboptimal annotation quality and imbalanced data distributions. Second, existing MSTI approaches predominantly operate as black boxes. While they can classify whether specific textual spans or image regions are sarcastic, they fail to articulate the underlying reasoning process, lacking interpretability.
To overcome these limitations, we curate the \textbf{MSTI-MAX} dataset, mitigating long-tailed distributions. More importantly, to break the prevailing black-box paradigm, we propose \textbf{GRASP}, which introduces Grounded CoT reasoning that leverages the structured text reasoning and preserves visual coordinates for target localization. It could effectively break the black-box nature of multimodal sarcasm, achieving advancements in both target identification accuracy and interpretability.



\begin{table}[t]
\centering
\caption{Statistics of MSTI~2.0 and MSTI-MAX datasets.}
\vspace{-1em}                     
\label{tab:dataset}
\begin{tabular}{ccccc}
\toprule
\textbf{Dataset} & \textbf{Split} & \textbf{Sarcasm} & \textbf{Non-sarcasm} & \textbf{Total} \\
\midrule
\multirow{4}{*}{\textbf{MSTI~2.0}} & Train & 3,500 & 46 & 3,546 \\
 & Val & 711 & 16 & 727 \\
 & Test & 729 & 13 & 742 \\
 & Total & 4,940 & 75 & 5,015 \\
\midrule
\multirow{4}{*}{\textbf{MSTI-MAX}} & Train & 3,500 & 3,046 & 6,546 \\
 & Val & 711 & 516 & 1,227 \\
 & Test & 729 & 513 & 1,242 \\
 & Total & 4,940 & 4,075 & 9,015 \\
\bottomrule
\end{tabular}
\vspace{-1em}
\end{table}
\begin{figure}[t]
    \centering
    \Description{Proportions of different datasets, 
    Fine-grained Sarcasm Modality Distribution of MSTI-MAX.}
    \includegraphics[width=0.48\textwidth]{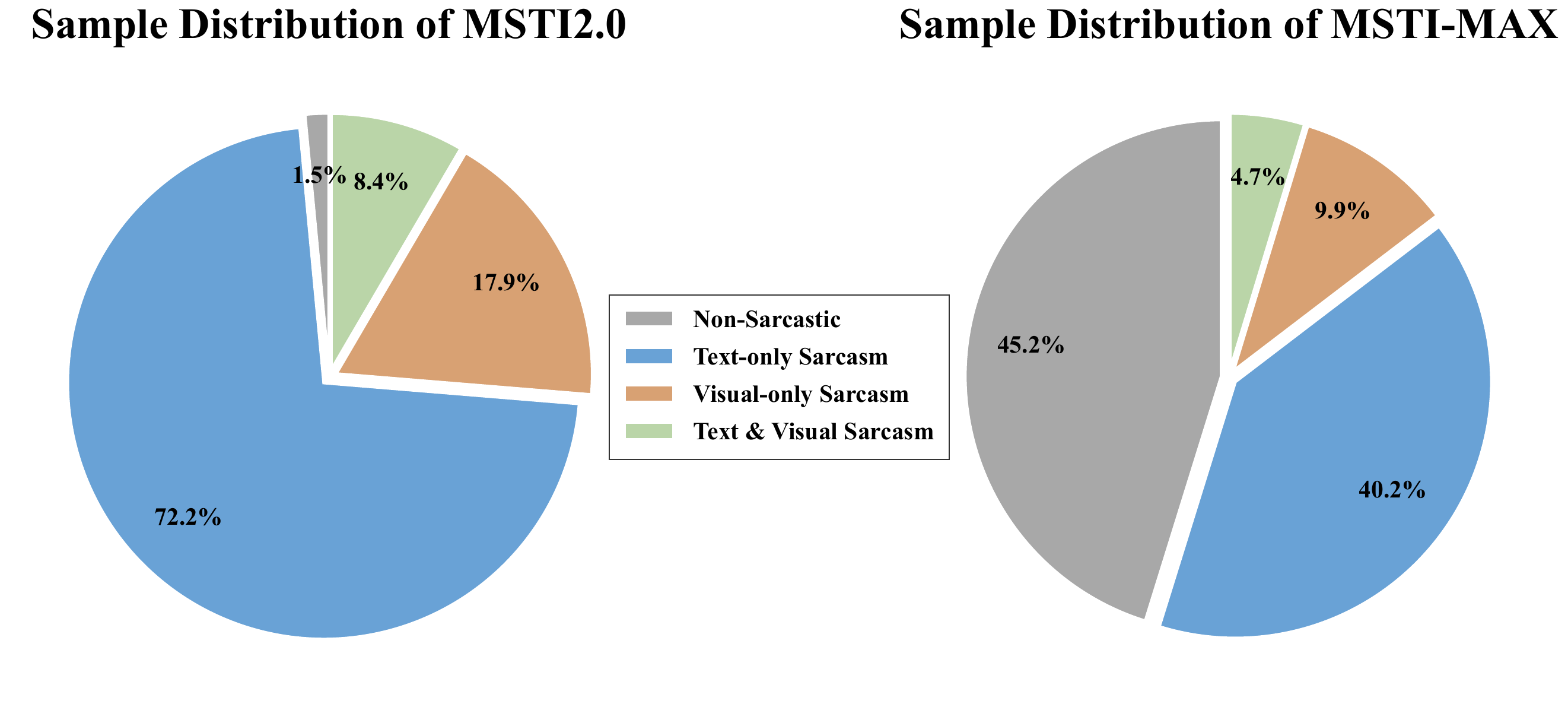}
    \vspace{-3em}                     
    \caption{Comparison of the fine-grained distribution of sarcasm targets across different modalities in the MSTI-MAX and MSTI~2.0 datasets. To reflect the diverse nature of real-world sarcastic expressions, the targets within each instance are categorized into three types: exclusively in the text (Text-only), exclusively in the visual modality (Visual-only), and present in both modalities (Text \& Visual).}
    \vspace{-1.5em}   
    \label{fig:data_fine}
\end{figure}

\begin{figure*}[t]
    \centering
    \Description{Overview of the GRASP training pipeline.}   
    \includegraphics[width=0.95\textwidth]{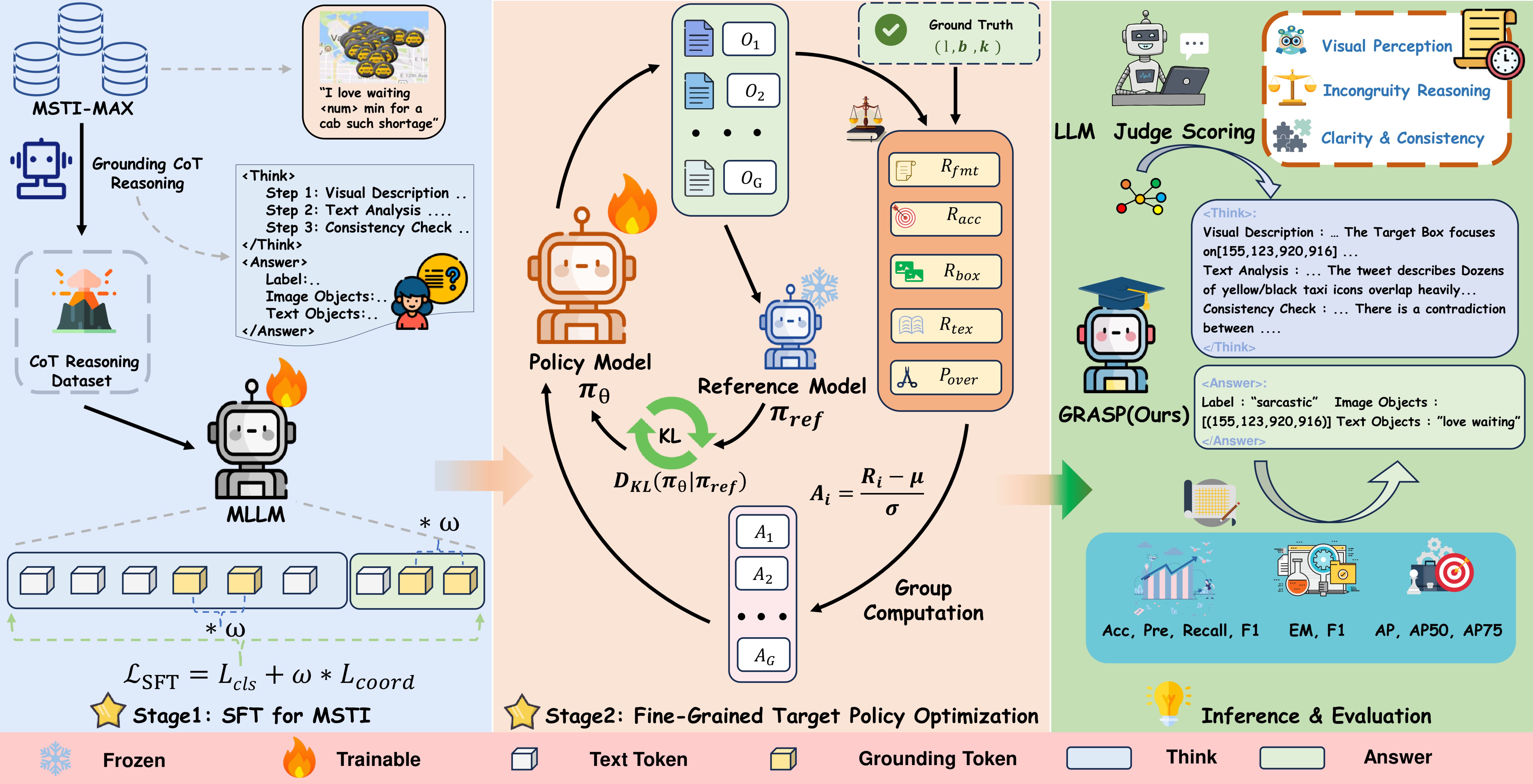}
    \caption{Overview of the GRASP training pipeline. The framework consists of two stages: (1)~Supervised Fine-Tuning (SFT) with coordinate-aware weighted loss employing structured grounded CoT reasoning; (2)~Fine-Grained Target Policy Optimization (FTPO) with multidimensional rewards, which further strengthens reasoning quality and visual grounding precision.}
    \vspace{-1.5em}
    \label{fig:overview}
\end{figure*}
\section{MSTI-MAX Dataset}
To mitigate data imbalance and enhance overall annotation quality, we construct the refined \textbf{MSTI-MAX} benchmark, building upon the original MSTI~2.0 dataset~\cite{chen2024cofipara}. The construction process addresses three critical aspects.

First, we observe a pronounced class imbalance in MSTI~2.0, as shown in Table~\ref{tab:dataset}. The overrepresentation of sarcastic samples not only inherently biases model training but also deviates from real-world communication scenarios, where non-sarcastic expressions naturally form the overwhelming majority of multimodal interactions. Training on such a skewed distribution leads to high false-positive rates in practical applications. To mitigate this discrepancy, we augment our dataset with high-quality, non-sarcastic image-text pairs sourced from established benchmarks for MSD, including a subset of data from training datasets of MMSD~\cite{cai2019multi} and MMSD~2.0~\cite{qin2023mmsd2}. The crucial augmentation aligns the data distribution more closely with the real world, thereby fostering training stability, reducing inductive bias, and ensuring robust generalization for practical deployments.
Second, the original dataset lacks bounding box annotations for key visual sarcasm targets in certain samples, which hinders fine-grained grounded reasoning. To address this, we manually review the samples labeled as sarcastic in MSTI~2.0 and supplement the missing visual annotations. To ensure the accuracy and standardization of the annotated data, we invite three domain experts in computer science to conduct independent annotations. All annotators possess solid analytical skills and extensive experience in data annotation. Prior to the formal annotation phase, they underwent unified training to align their understanding of the annotation guidelines and task requirements.
Furthermore, to ensure numerical stability and scale invariance during coordinate prediction, we standardize the geometric annotations by mapping all absolute pixel-level bounding boxes into a unified relative coordinate space of \([0, 1000]\).

Ultimately, MSTI-MAX is constructed to provide a balanced data distribution and high-quality, fine-grained annotations, establishing the solid foundation for explicit Grounded CoT reasoning in multimodal sarcasm target identification. The detailed statistics of both datasets are presented in Table~\ref{tab:dataset}, and the fine-grained sarcasm modality distribution is visualized in Figure~\ref{fig:data_fine}.

\vspace{-0.5em}
\section{Method}
In this section, we present the proposed GRASP framework for MSTI, and the complete training pipeline is shown in Figure~\ref{fig:overview}.
\vspace{-1em}
\subsection{Task Formulation}
Given a multimodal sarcasm dataset, $D=\{(V_i, T_i; l_i, b_i, k_i)\}_{i=1}^N$,
for the $i$-th sample, $V$ denotes the input image and $T$ denotes the input text.
$l \in \{0,1\}$ indicates whether the image-text pair is sarcastic.
If $l=1$, $b_i$ denotes the set of relative bounding boxes for visual anomalies that trigger sarcasm,
and $k$ denotes the textual trigger keywords.
Our goal is to generate the explicit reasoning $r$, the sarcasm label $\hat{l}$, and the multimodal sarcasm targets $\hat{b}, \hat{k}$.
\begin{equation}
    \mathcal{M}(V_i, T_i \mid \mathcal{T}_{stru}) \Rightarrow Y, \quad \text{where } Y = (r_i, a_i) ,
\end{equation}
where $\mathcal{T}_{stru}$ is the designed prompt template, $r$ is the explicitly generated chain-of-thought rationale in \texttt{<Think>} tags, and $a=(\hat{l}, \hat{b}, \hat{k})$ is the standardized final output in \texttt{<Answer>} tags.

\vspace{-1em}
\subsection{Grounded Chain-of-Thought Reasoning}

To achieve precise and interpretable multimodal sarcasm target identification, we introduce \textit{Grounded Chain-of-Thought (CoT) Reasoning}. Specifically, we employ an advanced Multimodal Large Language Model (MLLM), namely Qwen3.5-Plus~\cite{qwen35blog}, as the teacher model to synthesize high-quality reasoning data. 
Rather than directly jumping to conclusions, the model is mandated to first conduct a comprehensive analysis of the multimodal context. Building upon this holistic understanding, it performs grounded reasoning by explicitly ``Thinking with Images''. Given an image-text pair \((V, T)\) and its corresponding ground-truth labels \((l, k, b)\) in the training phase, we prompt the teacher model to generate structured reasoning paths via progressive multi-stage instructions:

\begin{enumerate}[leftmargin=13pt]
\item \textbf{Reasoning Stage}: The prompt guides the model to articulate its reasoning process within \texttt{<Think>...</Think>} tags, following three steps:
(i) \textit{Visual Description}, which extracts key objects and outputs their normalized coordinates;
(ii) \textit{Text Analysis}, which decodes the underlying meaning and tone of the textual modality;
(iii) \textit{Consistency Check}, which explicitly compares the visual and textual elements to evaluate semantic consistency.

\item \textbf{Decision Stage}: Following the reasoning process, the model outputs the final predictions within \texttt{<Answer>...</Answer>} tags. It includes the predicted label \(\hat{l}\), the textual target \(\hat{k}\), and the  visual target \(\hat{b}\) formatted as \([\hat{x}_\text{min}, \hat{y}_\text{min}, \hat{x}_\text{max}, \hat{y}_\text{max}]\).
\end{enumerate}

To ensure the quality and validity of the synthesized reasoning data, we invite three master’s students in computer science with professional annotation training to manually verify all samples. Guided by the ground-truth annotations, the generated reasoning of each sample is reviewed to ensure that the reasoning trajectories strictly adhered to our designed templates and logically entailed the correct standard answers. 
By deriving final predictions through this structured reasoning trajectory, our approach not only provides explicit rationales for enhanced interpretability but also establishes a seamless interface for reward computation in the subsequent reinforcement learning phase.
Detailed prompt templates can be found in the supplementary file.


    

\vspace{-1em}
\subsection{Dual-Stage Outcome-supervised Joint Optimization}
Inspired by the remarkable success of reinforcement learning from human feedback (RLHF) in aligning large language models~\cite{ouyang2022instructgpt, rafailov2023direct, deepseek2025r1}, we propose a dual-stage outcome-supervised joint optimization strategy tailored for our task.
In the first stage, we conduct Supervised Fine-Tuning (SFT) augmented with a \textbf{customized weighted cross-entropy loss} specifically designed for visual grounding. This stage equips the model with the fundamental capabilities required for the Multimodal Sarcasm Target Identification (MSTI) task, including adherence to the task paradigm, fine-grained target localization, and foundational cross-modal alignment.
Subsequently, to achieve more precise sarcasm target localization and deeper cross-modal understanding, \textbf{Fine-Grained Target Policy Optimization (FTPO)} is further introduced in the second stage. By leveraging multiple rewards optimization, FTPO can strengthen the model's target identification precision and its ability to perceive subtle cross-modal incongruence.


\subsubsection{Stage 1: SFT for MSTI}
During the first phase of training, our primary objective is to endow the model with fine-grained sarcasm target localization capabilities, activate cross-modal alignment, and elicit logical chain-of-thought reasoning. Standard MLLM instruction tuning typically applies an equal weight to all autoregressively generated tokens within the cross-entropy loss. 
Crucially, in multimodal sarcasm target identification, the precision of bounding box generation directly dictates the quality of visual grounding, which is paramount for comprehending complex multimodal semantics. Even if a sarcastic instance is correctly classified, inaccurate spatial coordinates indicate that the model fails to capture the true root cause of the visual sarcasm.

To allocate more optimization resources to spatial coordinate regression during backpropagation, we reformulate the training objective by introducing a \textit{Coordinate-aware Weighted Loss}  tailored for visual grounding, as the left part of Figure~\ref{fig:overview} shows. The asymmetric gradient feedback prevents the model from adopting spurious inference shortcuts that rely excessively on textual cues. Consequently, it not only improves localization accuracy but also implicitly encourages the model to focus on fine-grained visual details through context-based reasoning during the explicit ``\textit{Think}'' stage.
Given a multimodal input $X=(V,T)$ and target sequence $Y=\{y_1,y_2,\dots,y_N\}$, the refined autoregressive loss is:
\begin{equation}
\mathcal{L}_\text{SFT}= -\sum_{t=1}^N \omega_t \log P_\theta(y_t\mid X,Y_{<t}),
\end{equation}
where $\omega_t$ is a dynamic weight factor.
\begin{equation}
\label{eq:w_loss}
\omega_t = \begin{cases} 
\lambda_\text{coord}, & \text{if } y_t \in [x_\text{min},y_\text{min}, x_\text{max},y_\text{max}] \\ 
1, & \text{otherwise},  
\end{cases}
\end{equation}
where \(\lambda_\text{coord}\) serves as a scaling hyperparameter. For standard tokens, such as those constituting the reasoning rationales, general vocabulary, and classification labels, we maintain a base weight of \(\omega_t = 1\). Conversely, to explicitly emphasize the importance of precise visual grounding, tokens denoting relative spatial coordinates are assigned an amplified weight \(\omega_t = \lambda_\text{coord}\) and \(\lambda_\text{coord} > 1\).

\subsubsection{Stage 2: Fine-Grained Target Policy Optimization (FTPO)}

While the model acquires foundational capabilities for MSTI following the SFT stage, it often still struggles with imprecise target localization and reasoning-output misalignment when confronted with complex cross-modal incongruities. To further elevate the accuracy of sarcasm identification and the precision of visual grounding, we introduce an outcome-supervised reinforcement learning paradigm~\cite{lee2023rlaif}. 
Building upon GRPO~\cite{shao2024deepseekmath} and tailoring it to the unique characteristics of fine-grained target localization, we propose Fine-Grained Target Policy Optimization (FTPO). During training, for each image-text input pair \((V, T)\), FTPO circumvents the need for an external critic network. Instead, it samples a group of \(G\) candidate outputs \(\mathcal{O}=\{o_1,\dots,o_j, \dots, o_G\}\) directly from the current policy model \(\pi_\theta\). For each output \(o_j\), which comprises both the grounded Chain-of-Thought (CoT) rationale and the final answer, we compute an absolute reward \(R(o_j)\).
To ensure the fidelity and efficacy of the grounded Chain-of-Thought (CoT) reasoning for MSTI, we formulate a multi-dimensional reward mechanism. By imposing these comprehensive constraints, FTPO empowers the model to accurately discern subtle cross-modal incongruities within complex contexts, mitigate information redundancy, and achieve precise, traceable reasoning.

\begin{itemize}[leftmargin=10pt]
    \item \textbf{Format Compliance Reward (\(R^\text{fmt}\))}: Verifying whether the generated outputs strictly adhere to the predefined format in reasoning, specifically checking for valid \texttt{<Think>} and \texttt{<Answer>} tags. This reward is granted exclusively when all required structural components are intact.
    \item \textbf{Semantic Accuracy Reward (\(R^\text{acc}\))}: Evaluating the correctness of the sarcasm classification. This reward is assigned if the predicted label \(\hat{l}\) matches the ground truth \(l\).
     \item \textbf{Visual Grounding Reward (\(R^\text{box}\))}: Quantifying the spatial alignment by calculating the Intersection over Union (IoU) between the predicted bounding boxes \(\hat{b}\) and the ground truth \(b\). This explicitly compels the model to ground its visual attention on specific regions exhibiting cross-modal incongruity during the \texttt{<Think>} stage.
    \item \textbf{Text Exact Match Reward (\(R^\text{txt}\))}: Employing an Exact Match (EM) metric to quantify the alignment between the predicted textual trigger spans \(\hat{k}\) and the ground truth \(k_\text{gt}\).
    \item \textbf{Redundant Generation Penalty (\(P^\text{over}\))}: Imposing the penalty on the generation of excessive bounding boxes, overly verbose reasoning texts, or trivial full-sentence copying. This regularization effectively prevents the model from adopting spurious reward-hacking behaviors.
\end{itemize}
The total reward for any generated sequence $o_j$ is:
\begin{equation}
\label{eq:reward_beta}
R^\text{total}=\beta^1 R^\text{fmt} + \beta^2 R^\text{acc} + \beta^3 R^\text{box} + \beta^4 R^\text{txt} -\beta^5 P^\text{over},
\end{equation}
where \(\{\beta^k\}_{k=1}^5\) are hyperparameters to balance different rewards.

Subsequently, we calculate the group means and standard deviation of $\textbf{R}^{total} = \{R^{total}_1, \dots, R^{total}_G\}$, deriving the relative advantage for policy gradient updating.
\begin{equation}
\widetilde{R}^{total}_j = \frac{R^{total}(o_j)-mean(\textbf{R}^{total})}{std(\textbf{R}^{total})}.
\end{equation}
\begin{table*}[t]
\centering
\caption{Main results on the MSTI-MAX dataset. The best results in each column are highlighted in \textbf{bold}, and the second-best are \underline{underlined.} ``--'' indicates the metric is not applicable. ``*'' means the reproduced result.}
\vspace{-1em}
\label{tab:main}
\begin{tabular*}{\textwidth}{@{\extracolsep{\fill}} lccccccccc}
\toprule
\multirow{2}{*}{\textbf{Methods}} & \multicolumn{4}{c}{\textbf{Sarcasm Detection}} & \multicolumn{2}{c}{\textbf{Text Target}} & \multicolumn{3}{c}{\textbf{Visual Target}} \\
\cmidrule(lr){2-5} \cmidrule(lr){6-7} \cmidrule(lr){8-10}
 & \textbf{Acc} & \textbf{Precision} & \textbf{Recall} & \textbf{F1} & \textbf{EM} & \textbf{F1} & \textbf{AP} & \textbf{AP50} & \textbf{AP75} \\
\midrule
\rowcolor{gray!20}
\multicolumn{10}{l}{\textit{Text-only Baselines}} \\
BERT-base & 80.99 & 87.86 & 78.46 & 82.89 & 17.23 & 39.20 & -- & -- & -- \\
BERT-large & 81.59 & 85.98 & 79.97 & 82.87 & 21.74 & \underline{42.53} & -- & -- & -- \\
\midrule
\rowcolor{gray!20}
\multicolumn{10}{l}{\textit{Task-specific Multimodal Model}} \\
CofiPara & 80.20 & 73.52 & 85.32 & 78.98 & 49.52 & \textbf{48.74} & 9.79 & 23.18 & 8.00 \\
\midrule
\rowcolor{gray!20}
\multicolumn{10}{l}{\textit{Zero-shot Results of MLLMs with Large-scale Parameters}} \\
Gemini-3-Pro & \textbf{86.39} & \textbf{90.69} & 85.95 & \textbf{88.07} & 1.61 & 11.92 & 7.28 & 13.06 & 6.35 \\
Qwen3.5-Plus & 84.54 & 83.52 & 91.76 & \underline{87.45} & 3.62 & 12.73 & 9.34 & 13.68 & 8.70  \\
Qwen3.5-27B & \underline{84.78} & 87.09 & 86.97 & 87.03 & 4.59 & 14.69 & 27.29 & 32.27 & 26.97 \\
Qwen3-VL-32B & 72.22 & 75.07 & 78.88 & 76.92 & 11.19 & 10.51 & 23.10 & 27.64 & 22.57 \\
\midrule
\rowcolor{gray!20}
\multicolumn{10}{l}{\textit{Zero-shot Results of Lightweight MLLMs}} \\
Qwen3.5-4B & 67.71 & 72.97 & 71.47 & 72.21 & 10.63 & 10.96 & 15.38 & 18.86 & 14.98 \\
Qwen3.5-9B & 69.48 & 80.28 & 63.64 & 71.00 & 36.88 & 10.62 & 8.37 & 11.20 & 8.21 \\
Qwen3-VL-4B & 71.26 & 68.13 & \textbf{95.88} & 79.66 & 25.85 & 7.71 & 31.05 & 35.59 & 30.57 \\
Qwen3-VL-8B & 71.50 & 68.81 & \underline{94.10} & 79.49 & 4.43 & 10.63 & 29.57 & 34.08 & 29.07 \\

\midrule
\rowcolor{gray!40}
\multicolumn{10}{l}{\textit{Fine-tuned (Ours)}} \\
\textbf{GRASP} (Qwen3.5-4B) & 81.48 & 85.39 & 82.58 & 83.96 & 48.15 & 35.58 & 30.29 & 33.85 & 29.97 \\
\textbf{GRASP} (Qwen3.5-9B) & 83.74 & \underline{89.74} & 81.62 & 85.49 & \textbf{52.25} & 37.62 & \underline{42.51} & \underline{46.98} & \underline{41.78} \\
\textbf{GRASP} (Qwen3-VL-4B) & 78.50 & 77.11 & 90.12 & 83.11 & 43.16 & 35.04 & 41.31 & 45.84 & 40.74 \\
\textbf{GRASP} (Qwen3-VL-8B) & 84.06 & 85.73 & 87.38 & 86.55 & \underline{49.92} & 41.14 & \textbf{46.71} & \textbf{51.30} & \textbf{46.08} \\

\bottomrule
\end{tabular*}
\vspace{-1em}
\end{table*}

Using this relative advantage $\hat{A}^{total}_{j,t} = \widetilde{R}^{total}_j$, the model updates policy parameters $\theta$ via clipped importance sampling and KL divergence constraints,
driving convergence toward high-reward reasoning paths.
\begin{align}
&\mathcal{J}_\text{FTPO}(\theta) = \mathbb{E}\left[q\sim P(Q),\{o_j\}_{j=1}^G\sim\pi_{\theta_\text{old}}(O|q)\right] \notag\\
& \frac{1}{G}\sum_{j=1}^G\frac{1}{|o_j|}\sum_{t=1}^{|o_j|}\Bigg\{\min\Bigg[\frac{\pi_\theta(o_{j,t}|q,o_{j, <t})}{\pi_{\theta_\text{old}}(o_{j,t}|q,o_{j, <t})}\hat{A}^{total}_{j,t},\notag\\
& \mathrm{clip}\Bigg(\frac{\pi_\theta(o_{j,t}|q,o_{j, <t})}{\pi_{\theta_\text{old}}(o_{j,t}|q,o_{j, <t})},1-\varepsilon,1+\varepsilon\Bigg)\hat{A}^{total}_{j,t}\Bigg]
-\beta\mathbb{D}_\text{KL}\left[\pi_{\theta}||\pi_\text{ref}\right]\Bigg\}
\end{align}

\vspace{-1em}
\section{Experiments}
To evaluate the effectiveness and interpretability of the proposed GRASP framework on Multimodal Sarcasm Target Identification (MSTI) and Multimodal Sarcasm Detection (MSD), we conduct comprehensive experiments on our newly constructed MSTI-MAX dataset. This section details the experimental setup and addresses the following core Research Questions (RQs):
\begin{itemize}[leftmargin=10pt]
    \item \textbf{RQ1 (Superiority)}: How does GRASP perform on sarcasm detection and fine-grained target identification compared to state-of-the-art task-specific models and general-purpose Multimodal Large Language Models (MLLMs)?
    \item \textbf{RQ2 (Interpretability)}: Does the explicit Chain-of-Thought (CoT) reasoning generated by GRASP align with human cognitive logic, and how does it facilitate accurate MSTI?
    \item \textbf{RQ3 (Ablation \& Generalization)}: What are the individual contributions of the core components within GRASP, and how well does GRASP generalize across diverse foundational MLLMs?
\end{itemize}

\vspace{-1em}
\subsection{Experimental Setup}

\paragraph{Dataset and Evaluation Metrics} 
Experiments are conducted on our newly curated MSTI-MAX dataset. For MSD, we adopt Accuracy (Acc), Macro-F1, Precision, and Recall as metrics. For MSTI, following \cite{chen2024cofipara}, the evaluation is divided into two modalities: (1) For the identification of the textual target, we utilize the precision of the exact match (EM) \cite{joshi2018sarcasm} and the F1 score \cite{joshi-2019-overview_sarcastic}; (2) For visual target identification, we employ standard COCO-style Average Precision (AP) metrics \cite{Lin-COCO}, specifically AP, AP50, and AP75. Details of metrics can be found in the supplementary file.

\paragraph{Baselines}
To comprehensively evaluate our method, we select a diverse set of baselines spanning three categories: (1) text-only models (e.g., BERT~\cite{devlin2019bert}); (2) the state-of-the-art (SOTA) task-specific multimodal model, CofiPara~\cite{chen2024cofipara}; and (3) a range of prominent Multimodal Large Language Models (MLLMs), including Gemini-3-Pro, as well as the Qwen3-VL~\cite{qwen2025qwen3}, Qwen3.5~\cite{qwen35blog}, and InternVL3.5~\cite{wang2025internvl35} series. 
For CofiPara~\cite{chen2024cofipara}, we reproduce the results using their officially released source code. Since the ``contrastive rationales'' required by their method were not publicly available, we utilize Qwen2.5-VL~\cite{bai2025qwen25vl} to regenerate them. Our reproduce results on the original MSTI~2.0 dataset align consistently with the performance reported in their paper, ensuring a fair and rigorous comparison.

\paragraph{Implementation Details}
By default, GRASP adopts the QwenVL series as the foundation. During SFT, we use Low-Rank Adaptation (LoRA) with \(r=4\) and set the bounding box coordinate weight \(\lambda_\text{coord}\) in Eq. \ref{eq:w_loss} to 10, and our experimental search shows this balances precise visual grounding and reasoning text fluency. For FTPO, the reinforcement learning sampling group size is G=4, and hyperparameters, including ${\beta^1, \beta^2, \beta^3, \beta^4, \beta^5}$ in Eq. \ref{eq:reward_beta}, are set to 0.05, 0.15, 0.4, 0.4, 0.3. This uneven distribution prioritizes core abilities, while small weights for format $\beta^1$ suffice to maintain structural constraints learned in SFT. Details can be found in the supplementary.

\vspace{-1em}
\subsection{Main Results}


To answer RQ1, we conduct comprehensive experiments across diverse MLLMs with varying parameter scales, as shown in Table~\ref{tab:main}. 

\begin{enumerate}[leftmargin=12pt]
    \item \textbf{Superior Performance in Both MSD and MSTI}: 
    The results demonstrate that GRASP significantly outperforms current State-of-the-Art (SOTA) models in visual sarcasm target localization, while achieving highly competitive performance in both MSD and textual target identification. 
    It confirms that GRASP's integration of fine-grained visual grounding with structured CoT reasoning effectively tackles the MSTI task by fostering deeper cross-modal semantic alignment. 
    The reproduced CofiPara experiences a substantial performance degradation on MSTI-MAX. We attribute this primarily to the significant class distribution shift, transitioning from the highly imbalanced MSTI~2.0 to MSTI-MAX (from 1.5\%  to 45.2\% no-sarcasm samples).
    \item \textbf{Necessity of Task-Specific Fine-Tuning}: 
    While massive proprietary models, like the Gemini-3-Pro and Qwen3.5-Plus, achieve outstanding performance on simple MSD tasks by leveraging their strong general capabilities, they struggle significantly with fine-grained target localization in the zero-shot setting. These general-purpose models often fail to capture the subtle multimodal nuances of sarcasm, frequently generating inaccurate visual bounding boxes. In contrast, our GRASP framework, despite utilizing parameter-efficient fine-tuning on much smaller base models, achieves profound task alignment. It outperforms these massive general-purpose models by a significant margin, underscoring the effectiveness of GRASP in MSTI.
    \item \textbf{Key Empirical Observations}:
    During our evaluation, we observe several notable findings. First, while scaling up model parameters generally benefits the relatively simpler binary MSD task, this scaling law does not strictly apply to the more complex, fine-grained MSTI task. For instance, despite its massive scale, the Gemini-3-Pro model exhibits suboptimal performance in target localization compared to the lightweight Qwen3-VL-4B.
    Second, newer model iterations do not universally guarantee performance gains on specific multimodal tasks. In our experiments, the Qwen3-VL series significantly outperforms the newer Qwen3.5 series. We attribute this discrepancy to the focus of their respective pre-training phases. Qwen3-VL underwent extensive visual-centric enhancements, e.g., upgraded visual encoding and fine-grained recognition, whereas Qwen3.5 prioritized optimization in other domains, inadvertently leading to catastrophic forgetting of certain fine-grained visual capabilities. 
    Nevertheless, despite the inherently weak zero-shot performance of Qwen3.5 on the MSTI task, integrating it into the GRASP framework yields substantial performance improvements. This effectively mitigates the catastrophic forgetting and further validates the robustness and efficacy of our approach.  
\end{enumerate}


\subsection{Evaluation of Reasoning Quality}
\begin{figure}[h]
    \centering
    \Description{LLM-as-a-Judge evaluation results.}   
    \includegraphics[width=0.75\linewidth]{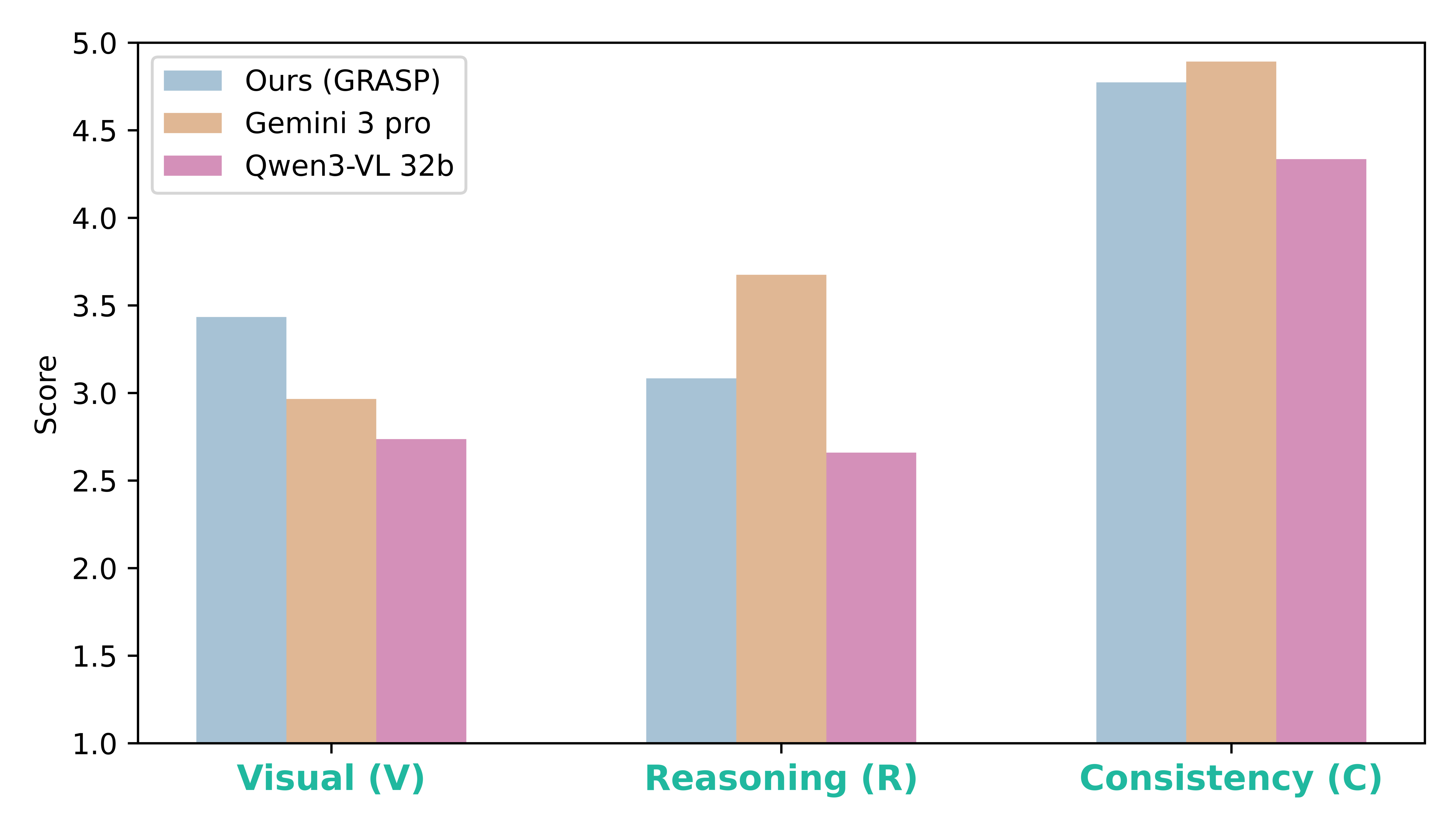}
    \vspace{-1.5em}
    \caption{LLM-as-a-Judge evaluation results. Higher values indicate better performance on a 1--5 scale.}
    \vspace{-1.2em}
    \label{fig:radar}
\end{figure}


To address RQ2, we move beyond standard outcome-oriented metrics and adopt the LLM-as-a-Judge paradigm~\cite{zheng2023judging,liu2023geval}. It allows us to evaluate whether the model genuinely acquires the cognitive capacity to comprehend cross-modal sarcasm, rather than merely overfitting to surface-level dataset biases. Specifically, we employ Qwen3.5-Plus as the evaluator to conduct blind, quantitative scoring (on a scale of 1 to 5) on the reasoning rationales generated by various models. The evaluation spans three core dimensions, including Visual Perception (V), Reasoning Ability (R), and Logical Consistency (C). 
As illustrated in Figure~\ref{fig:radar}, GRASP (Qwen3-VL-8B) achieves comparable, if not superior, results relative to the massive Gemini-3-Pro model. Notably, it excels in Visual Perception and Logical Consistency, underscoring the high quality and interpretability of its explicit chain-of-thought reasoning. Details of the evaluation can be found in the supplementary.
\vspace{-1em}
\begin{table*}[t]
\centering
\caption{Ablation study of the proposed components based on Qwen3-VL-4B and generalization performance across various foundational MLLMs on the MSTI-MAX dataset. The best results within each group are highlighted in \textbf{bold}.}
\vspace{-1em}
\label{tab:ablation}
\begin{tabular*}{\textwidth}{@{\extracolsep{\fill}} l cccc cc ccc}
\toprule
\multirow{2}{*}{\textbf{Variants}} & \multicolumn{4}{c}{\textbf{Sarcasm Detection}} & \multicolumn{2}{c}{\textbf{Text Target}} & \multicolumn{3}{c}{\textbf{Visual Target}} \\
\cmidrule(lr){2-5} \cmidrule(lr){6-7} \cmidrule(lr){8-10}
 & \textbf{Acc} & \textbf{Precision} & \textbf{Recall} & \textbf{F1} & \textbf{EM} & \textbf{F1} & \textbf{AP} & \textbf{AP50} & \textbf{AP75} \\
\midrule
\rowcolor{gray!20}
\multicolumn{10}{l}{\textit{Ablation Study of GRASP Based on Qwen3-VL-4B}} \\
GRASP (Qwen3-VL-4B) & \textbf{78.50} & 77.11 & \textbf{90.12} & \textbf{83.11} & 43.16 & \textbf{35.04} & \textbf{41.31} & \textbf{45.84} & \textbf{40.74} \\
\quad w/o FTPO (SFT only) & 73.26 & \textbf{83.02} & 68.44 & 75.03 & \textbf{46.45} & 28.88 & 34.21 & 38.25 & 33.70 \\
\quad w/o SFT (FTPO only) & 72.71 & 77.69 & 75.03 & 76.34 & 45.11 & 7.62 & 12.94 & 17.24 & 12.58 \\
\quad w/o Bbox-Weight & 72.79 & 75.16 & 80.11 & 77.56 & 41.14 & 29.39 & 30.03 & 33.70 & 29.38 \\

\midrule
\rowcolor{gray!40}
\multicolumn{10}{l}{\textit{Generalization of GRASP on Various Foundation MLLMs}} \\
InternVL3.5-4B (zero-shot) & 65.38 & 74.07 & 63.10 & 68.15 & 21.01 & 8.93 & 5.19 & 7.72 & 4.50 \\
InternVL3.5-8B (zero-shot) & 68.84 & 75.91 & 68.72 & 72.14 & 11.59 & 11.32 & 17.86 & 21.22 & 16.87 \\
\textbf{GRASP} (InternVL3.5-4B) & 80.68 & 83.72 & 83.26 & 83.49 & 49.19 & 38.99 & 41.05 & 45.81 & 40.44 \\
\textbf{GRASP} (InternVL3.5-8B) & 81.80 & 83.22 & 86.42 & 84.79 & \textbf{50.24} & 34.88 & 42.95 & 47.40 & 42.22 \\
\textbf{GRASP} (Qwen3-VL-8B) & \textbf{84.06} & \textbf{85.73} & \textbf{87.38} & \textbf{86.55} & 49.92 & \textbf{41.14} & \textbf{46.71} & \textbf{51.30} & \textbf{46.08} \\
\bottomrule
\end{tabular*}
\vspace{-1em}
\end{table*}
\subsection{Ablation Study}

To answer RQ3, we conduct ablation experiments on the key components of the GRASP framework, as shown in the upper part of Table \ref{tab:ablation}.
\textbf{Without Supervised Fine-Tuning (w/o SFT)}: Bypassing SFT and directly applying FTPO severely hinders convergence. The model produces frequent formatting errors that trigger heavy penalties in RL, drastically reducing classification and localization accuracy. It confirms that SFT is indispensable, that is, it provides task-specific heuristics and high-quality policy initialization, which are critical for effective subsequent reinforcement learning.
\textbf{Without Fine-Grained Target Policy Optimization (w/o FTPO)}: Removing the reward-guided RL stage and using only SFT causes significant performance degradation. Although the SFT-only model learns the CoT structure, it often generates flawed reasoning inconsistent with final predictions. It demonstrates the necessity of FTPO. The multi-dimensional rewards encourage robust cross-modal alignment, moving beyond format imitation to real multimodal reasoning.
\textbf{Without Bounding Box Weighted Loss (w/o Bbox-Weight)}: Using standard unweighted cross-entropy loss \(\lambda_{box}=1\) degrades performance across all MSTI metrics. It shows that without emphasizing coordinate tokens, the model cannot prioritize fine-grained visual grounding, weakening its ability to locate fine-grained sarcasm targets accurately.

\vspace{-1em}
\subsection{Generalization of GRASP}
\begin{table}[t]
\centering
\caption{Generalization results on the MMSD~2.0 benchmark. Results of baselines are taken from GDCNet~\cite{zhang2026gdcnet-mmsd2}.}
\vspace{-1em}
\label{tab:mmsd2}
\begin{tabular}{lcccc}
\toprule
\textbf{Method} & \textbf{Acc} & \textbf{Precision}& \textbf{Recall} & \textbf{F1} \\
\midrule
LLaVA (Zero-Shot) & 51.06 & 40.09 & 46.40 & 43.02 \\
Qwen-VL (Zero-Shot) & 40.63 & 32.44 & 35.53 & 33.63 \\
GPT-4o (Zero-Shot) & 71.07 & \textbf{79.52} & 71.07 & 70.24 \\
LLaVA (CoT) & 48.69 & 40.93 & 65.17 & 50.28 \\
Qwen-VL (CoT) & 58.86 & 56.82 & 58.67 & 57.26 \\
GPT-4o (CoT) & 74.26 & 65.81 & 72.68 & 68.92 \\
\midrule
Qwen3-VL-8B(Zero-Shot) & 62.72 & 54.05 & \textbf{89.49} & 67.39 \\
\textbf{GRASP }(Qwen3-VL-8B) & \textbf{78.33} & 73.09 & 78.59 & \textbf{75.74} \\
\bottomrule
\end{tabular}
\vspace{-2em}
\end{table}

To assess the out-of-domain generalization capabilities of our approach, we directly test the GRASP model, trained exclusively on MSTI-MAX, on MMSD~2.0~\cite{qin2023mmsd2}, a standard benchmark for binary multimodal sarcasm detection. As detailed in Table~\ref{tab:mmsd2}, despite MMSD~2.0 requiring only coarse-grained binary labels, GRASP consistently outperforms existing baselines. The remarkable performance underscores the robust transferability and broad applicability of our method.
Furthermore, to validate the architectural robustness of GRASP, we substitute the default Qwen-VL backbone with InternVL~\cite{wang2025internvl35}, another advanced MLLM. As shown in the lower section of Table~\ref{tab:ablation}, the InternVL-equipped GRASP achieves excellent MSTI performance, on par with its Qwen-VL counterpart across all metrics. It confirms that GRASP serves as the generalizable framework, seamlessly adaptable to diverse MLLMs.


\subsection{Case Study}
\begin{figure}[t]
    \centering
    \Description{A case study showing the complete GRASP reasoning and target identification process on a ride-hailing app sarcasm example.}
    \includegraphics[width=\linewidth]{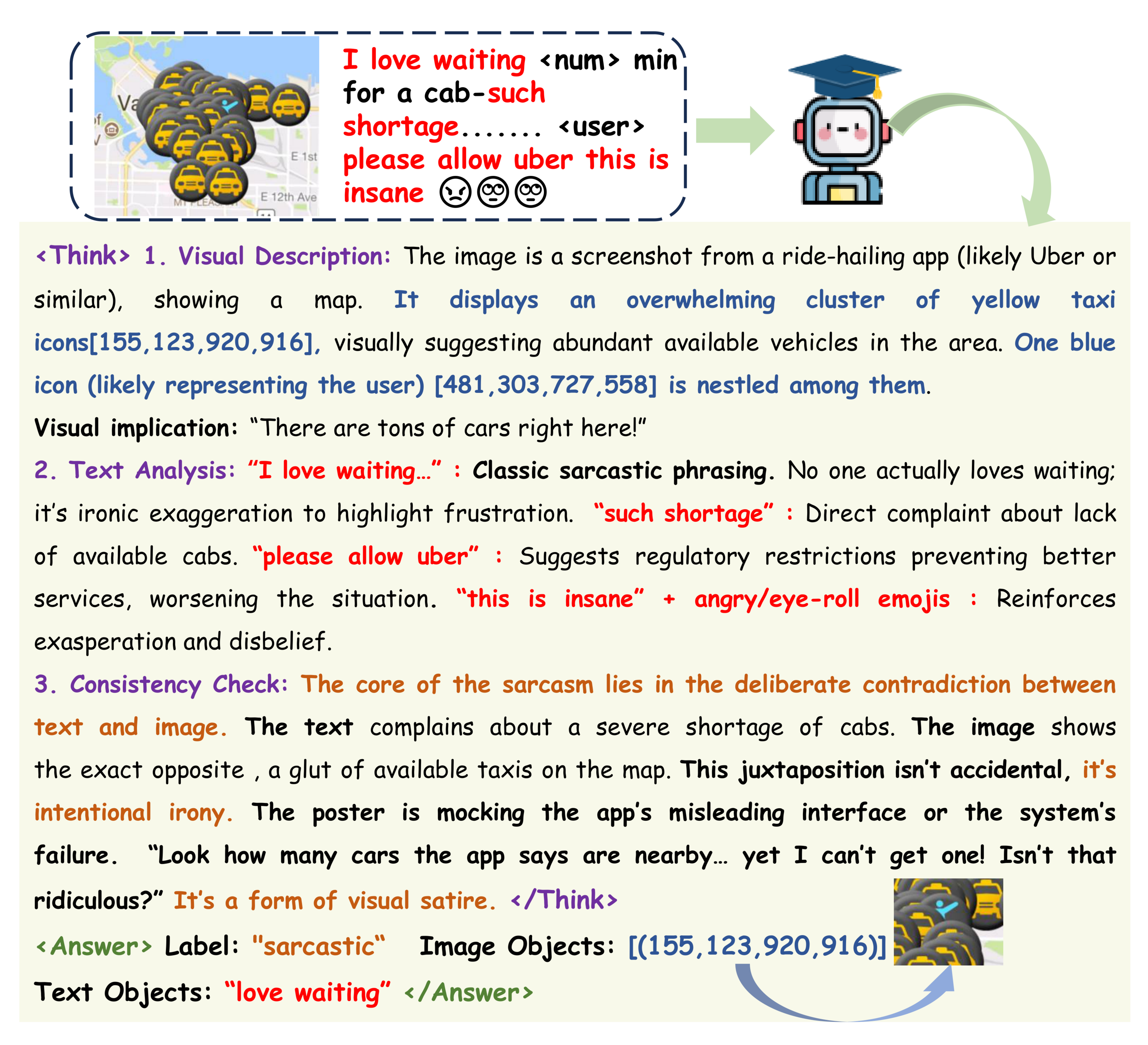}
    \vspace{-2.5em}
    \caption{A case study of GRASP on the MSTI task.}
    \label{fig:case}
    \vspace{-2em}
\end{figure}

To qualitatively illustrate how GRASP performs fine-grained sarcasm target identification, we present a representative example in Figure~\ref{fig:case}. 
Given a multimodal input, GRASP executes its Grounded CoT reasoning within the \texttt{<Think> </Think>} tags, which encompasses three key phases: visual description, textual analysis, and cross-modal consistency check. Guided by this structured rationale, GRASP first localizes the visual sarcasm target using precise bounding box coordinates and identifies the text sarcasm target words. Subsequently, it detects the core cross-modal contradiction by explicitly aligning the semantic cues from both modalities, ultimately yielding the final prediction within the \texttt{<Answer> </Answer>} tags. 
The case study demonstrates that GRASP can precisely pinpoint fine-grained sarcasm targets across both visual and textual modalities, further validating its practical effectiveness on MSTI.

\section{Ethics Statement}
All datasets utilized and reconstructed in this research for MSTI-MAX are derived from publicly available sources and strictly intended for academic purposes. 
We ensure that all instances are anonymized, with no personally identifiable information or social media user metadata included. 
Given the inherent nature of sarcasm, some data samples may contain cynical or mocking tones. However, we have taken measures to exclude severe hate speech, explicit toxicity, and discriminatory content.

\vspace{-1em}
\section{Conclusion}
In this paper, we present \textbf{GRASP}, a novel framework designed to address the challenges of fine-grained target localization and inadequate explainability for MSTI. To facilitate this research, we first introduce MSTI-MAX, a meticulously reconstructed benchmark featuring balanced distributions, fine-grained target annotations, i.e., visual bounding boxes and textual target words, and detailed reasoning rationales. Building upon this, GRASP leverages Grounded CoT reasoning to explicitly articulate the multimodal rationale behind sarcastic intents. Furthermore, we propose an innovative dual-stage optimization strategy. It begins with SFT utilizing a customized coordinate-aware weighted loss for initial task alignment, followed by FTPO with multiple rewards to iteratively refine both target localization and multimodal reasoning.
Extensive experiments demonstrate that GRASP achieves superior performance in both sarcasm detection and fine-grained target localization, while exhibiting strong  generalization and robustness. Notably, our pioneering LLM-as-a-Judge evaluation confirms that GRASP effectively captures the complex cross-modal incongruities inherent in sarcasm. Ultimately, this work establishes that integrating explicit visual grounding with structured CoT reasoning is the effective paradigm for complex multimodal tasks. We will explore more efficient training paradigms and broader contextual modeling to further advance multimodal sarcasm understanding and reasoning in future work.

\bibliographystyle{ACM-Reference-Format}
\bibliography{references}
\appendix
\section{Prompt Templates}
\label{sec:prompts}

The prompts used in this study are designed to guide the model through various reasoning and generation stages. The following is a detailed description of their roles: 

\begin{itemize}
  \item \textbf{Figure~\ref{fig:distill_prompt}} shows the prompt used to distill high-quality CoT trajectories from a powerful teacher model during the Supervised Fine-Tuning (SFT) stage. It standardizes the data generation process, ensuring that the reasoning chains explicitly anchor visual regions and strictly align with the ground-truth targets.
  
  \item \textbf{Figure~\ref{fig:infer_prompt}} presents the inference prompt designed to guide the model's Grounding CoT reasoning. It prompts the model to extract visual anchors and analyze cross-modal incongruities within the \texttt{<Think></Think>} module, before predicting the final classification label and localizing fine-grained targets in the \texttt{<Answer></Answer>} module.
  
  \item \textbf{Figure~\ref{fig:judge_prompt}} illustrates the prompt used for the LLM-as-a-Judge evaluation. It is designed to quantitatively measure the quality of the generated reasoning chains across three specific dimensions: \textit{Visual Perception}, \textit{Incongruity Reasoning}, and \textit{Logical Consistency}.
\end{itemize}
\section{Implementation Details}
\label{sec:impl_details}

\subsection{Training Hyperparameters}

Table~\ref{tab:hyperparams} summarizes the key hyperparameters for both the SFT and FTPO (Fine-Grained Target Policy Optimization) training stages.

\begin{table}[h]
\centering
\caption{Training hyperparameters for SFT and FTPO stages.}
\label{tab:hyperparams}
\begin{tabular}{lcc}
\toprule
\textbf{Hyperparameter} & \textbf{SFT} & \textbf{FTPO} \\
\midrule
LoRA Rank / Alpha       & 4 / 16       & 8 / 32        \\
Learning Rate           & 3.0e-5       & 1.0e-5        \\
Warmup                  & 80 steps     & 5\% ratio     \\
Num Epochs              & 2            & 1             \\
Effective Batch Size    & 8            & 16            \\
Max Sequence Length     & 2048         & 2048          \\
Coordinate Weight $\lambda_\text{coord}$ & 10        & --           \\
Precision               & bf16         & bf16          \\
\bottomrule
\end{tabular}
\end{table}

\section{Detailed Evaluation Metrics}
\label{sec:metrics}

Our evaluation framework covers two tasks: Multimodal Sarcasm Detection (MSD) and Multimodal Sarcasm Target Identification (MSTI). We detail each metric below.

\subsection{MSD Metrics}

For the binary sarcasm detection task, we adopt four standard classification metrics:

\begin{itemize}[nosep]
  \item \textbf{Accuracy (Acc)}: The proportion of correctly classified samples over the total number of samples:
  \begin{equation}
    \text{Acc} = \frac{TP + TN}{TP + TN + FP + FN}
  \end{equation}

  \item \textbf{Precision}: The fraction of predicted sarcastic samples that are truly sarcastic:
  \begin{equation}
    \text{Precision} = \frac{TP}{TP + FP}
  \end{equation}

  \item \textbf{Recall}: The fraction of truly sarcastic samples that are correctly identified:
  \begin{equation}
    \text{Recall} = \frac{TP}{TP + FN}
  \end{equation}

  \item \textbf{Macro-F1}: The unweighted average of the F1 scores computed independently for each class (sarcastic and not sarcastic), which accounts for class imbalance:
  \begin{equation}
    \text{Macro-F1} = \frac{1}{|C|}\sum_{c \in C} \frac{2 \cdot \text{Precision}_c \cdot \text{Recall}_c}{\text{Precision}_c + \text{Recall}_c}
  \end{equation}
\end{itemize}

\subsection{MSTI Metrics -- Textual Target Identification}

We evaluate textual target identification with two token-level metrics:

\begin{itemize}[nosep]
  \item \textbf{Exact Match (EM) Accuracy}: A strict metric that counts a prediction as correct only if the predicted text target \emph{exactly} matches the ground-truth string. For a dataset of $N$ samples:
  \begin{equation}
    \text{EM} = \frac{1}{N} \sum_{i=1}^{N} A_i, \quad \text{where } A_i = 
    \begin{cases} 
      1, & \text{if } \hat{k}_i = k_i \\ 
      0, & \text{otherwise} 
    \end{cases}
  \end{equation}
  where $\hat{k}_i$ is the predicted text target and $k_i$ is the ground-truth target for sample $i$.

  \item \textbf{F1 Score}: A token-level overlap metric. The predicted and ground-truth text targets are tokenized into sets of tokens. Precision and recall are computed based on the overlap, and F1 is their harmonic mean. This relaxes the strict EM criterion by giving partial credit for partially correct predictions:
  \begin{equation}
    P = \frac{|\hat{T} \cap T^*|}{|\hat{T}|}, \quad R = \frac{|\hat{T} \cap T^*|}{|T^*|}, \quad \text{F1} = \frac{2PR}{P+R}
  \end{equation}
  where $\hat{T}$ and $T^*$ denote the predicted and ground-truth token sets, respectively.
\end{itemize}

\subsection{MSTI Metrics -- Visual Target Identification}

For visual target identification, we adopt the standard COCO-style Average Precision (AP) metrics. A predicted bounding box $\hat{b}$ is matched to a ground-truth box $b$ using Intersection over Union (IoU):
\begin{equation}
  \text{IoU}(\hat{b}, b) = \frac{|\hat{b} \cap b|}{|\hat{b} \cup b|}
\end{equation}

We report three AP variants:
\begin{itemize}[nosep]
  \item \textbf{AP}: The primary COCO metric, computed as the mean AP averaged over 10 IoU thresholds from 0.50 to 0.95 with a step size of 0.05. This provides a comprehensive assessment across different localization strictness levels.

  \item \textbf{AP50}: AP at IoU threshold 0.50. A predicted box is considered correct if $\text{IoU} \geq 0.50$, offering a more lenient localization criterion.

  \item \textbf{AP75}: AP at IoU threshold 0.75. This stricter threshold demands more precise spatial localization of the sarcasm target.
\end{itemize}

The precision-recall curve is computed by ranking all detections by confidence score and sweeping across recall levels. AP is then calculated as the area under the interpolated precision-recall curve.

\section{Hyperparameter Selection}
\label{sec:hyperparam_selection}

To better justify the final hyperparameter configuration adopted in GRASP, we provide additional comparative analyses for the key hyperparameters used in the SFT and FTPO stages. We focus on two design choices: the coordinate-token weight $\lambda_\text{coord}$ in the weighted SFT loss and the reward coefficients $(\beta^1,\beta^2,\beta^3,\beta^4,\beta^5)$ in FTPO.

\subsection{Effect of Coordinate Weight \texorpdfstring{$\lambda_\text{coord}$}{lambda\_coord}}
\label{sec:add_exp}

In the main paper, we report the comparison between $\lambda_\text{coord}=1$ (i.e., w/o Bbox-Weight, standard unweighted loss) and $\lambda_\text{coord}=10$ (the selected setting used in the main experiments). Here, we extend that comparison to a wider range of weights by additionally including $\lambda_\text{coord}=5$ and $\lambda_\text{coord}=15$, as shown in Table~\ref{tab:coord_weight}.

\begin{table}[h]
\centering
\caption{Effect of the coordinate weight $\lambda_\text{coord}$ in the SFT stage on GRASP (Qwen3-VL-4B). Results for $\lambda_\text{coord} \in \{1, 10\}$ are reported in the main paper; the additional settings $\lambda_\text{coord} \in \{5, 15\}$ are supplemented here.}
\label{tab:coord_weight}
\resizebox{\columnwidth}{!}{%
\begin{tabular}{c cccc cc ccc}
\toprule
\multirow{2}{*}{$\lambda_\text{coord}$} & \multicolumn{4}{c}{\textbf{Sarcasm Detection}} & \multicolumn{2}{c}{\textbf{Text Target}} & \multicolumn{3}{c}{\textbf{Visual Target}} \\
\cmidrule(lr){2-5} \cmidrule(lr){6-7} \cmidrule(lr){8-10}
 & \textbf{Acc} & \textbf{Prec.} & \textbf{Rec.} & \textbf{F1} & \textbf{EM} & \textbf{F1} & \textbf{AP} & \textbf{AP50} & \textbf{AP75} \\
\midrule
1  & 72.79 & 75.16 & 80.11 & 77.56 & 41.14 & 29.39 & 30.03 & 33.70 & 29.38 \\
5  & 77.54 & \textbf{82.61} & 78.19 & 80.34 & \textbf{46.14} & 30.06 & 37.62 & 41.82 & 37.06 \\
\textbf{10} & \textbf{78.50} & 77.11 & \textbf{90.12} & \textbf{83.11} & 43.16 & \textbf{35.04} & \textbf{41.31} & \textbf{45.84} & \textbf{40.74} \\
15  & 69.00 & 67.77 & 89.99 & 77.31 & 46.38 & 11.90 & 22.92 & 25.39 & 22.70 \\
\bottomrule
\end{tabular}%
}
\end{table}

As shown in Table~\ref{tab:coord_weight}, increasing $\lambda_\text{coord}$ from 1 to 5 yields consistent improvements across nearly all metrics, particularly in sarcasm detection F1 (+2.78) and visual target AP (+7.59). Further increasing to $\lambda_\text{coord}=10$ brings additional gains, especially in recall (+11.93 over $\lambda_\text{coord}=5$) and text target F1 (+4.98), achieving the best overall balance. However, when the weight is pushed excessively high to $\lambda_\text{coord}=15$, we observe a severe performance degradation across all tasks. Notably, the textual target F1 score crashes to 11.90, and the visual target AP drops sharply to 22.92. This shows that appropriately increasing the loss weight for the target box helps the model better focus on image regions related to sarcasm, thus improving the accuracy of sarcasm target identification. However, an excessively large coordinate weight can damage the joint learning objective, causing the model to over-emphasize coordinate label generation at the cost of overall semantic understanding.  Accordingly, $\lambda_\text{coord}=10$ is selected for the main experiments as the most balanced choice.

\subsection{Effect of FTPO Reward Weight Allocation}
\label{sec:reward_weight}

In the main paper, the FTPO stage adopts the reward weights
\(
(\beta^1,\beta^2,\beta^3,\beta^4,\beta^5)=(0.05, 0.15, 0.40, 0.40, 0.30)
\),
which assign larger coefficients to the two grounding-oriented rewards, namely visual localization and textual target matching. To further verify the rationality of this design, we compare the selected reward allocation against an equal-weight baseline, an accuracy-oriented variant, and a format-enhanced variant. The comparison results are summarized in Table~\ref{tab:reward_weight}.

\begin{table}[h]
\centering
\setlength{\tabcolsep}{2.5pt}
\scriptsize
\caption{Effect of FTPO reward weight allocation on GRASP (Qwen3-VL-4B). Larger weights on the grounding-oriented rewards $R^\text{box}$ and $R^\text{txt}$ yield the best overall trade-off across MSD and MSTI metrics.}
\label{tab:reward_weight}
\resizebox{\columnwidth}{!}{%
\begin{tabular}{l cccc cc ccc}
\toprule
\multirow{2}{*}{\textbf{Setting}} & \multicolumn{4}{c}{\textbf{Sarcasm Detection}} & \multicolumn{2}{c}{\textbf{Text Target}} & \multicolumn{3}{c}{\textbf{Visual Target}} \\
\cmidrule(lr){2-5} \cmidrule(lr){6-7} \cmidrule(lr){8-10}
& \textbf{Acc} & \textbf{Prec.} & \textbf{Rec.} & \textbf{F1} & \textbf{EM} & \textbf{F1} & \textbf{AP} & \textbf{AP50} & \textbf{AP75} \\
\midrule
\textbf{GRASP} & 78.50 & 77.11 & \textbf{90.12} & \textbf{83.11} & 43.16 & \textbf{35.04} & \textbf{41.31} & \textbf{45.84} & \textbf{40.74} \\
Equal-weight & \textbf{78.66} & \textbf{79.67} & 85.46 & 82.46 & \textbf{43.32} & 32.66 & 39.81 & 43.83 & 39.36 \\
Acc-heavy & 75.12 & 77.78 & 80.66 & 79.19 & 42.67 & 31.20 & 36.46 & 40.30 & 35.85 \\
Format-enhanced & 76.73 & 76.51 & 87.11 & 81.46 & 40.18 & 32.29 & 39.22 & 43.33 & 38.88 \\
\bottomrule
\end{tabular}%
}
\vspace{2pt}
\parbox{\columnwidth}{\scriptsize\textit{Weight settings:} \textbf{GRASP} $(0.05, 0.15, 0.40, 0.40, 0.30)$; \textbf{Equal-weight} $(0.25, 0.25, 0.25, 0.25, 0.30)$; \textbf{Acc-heavy} $(0.05, 0.45, 0.25, 0.25, 0.30)$; \textbf{Format-enhanced} $(0.20, 0.20, 0.30, 0.30, 0.30)$.}
\end{table}

Table~\ref{tab:reward_weight} shows that the proposed weight allocation achieves the best overall performance balance. 
Although the equal-weight baseline yields marginally higher accuracy, precision, and EM, it is consistently inferior to the selected allocation on the more task-critical metrics, including sarcasm detection F1, text-target F1, and all three visual grounding metrics. 
It suggests that uniformly distributing reward mass is suboptimal for MSTI, where high-quality reasoning must jointly support fine-grained textual and spatial localization.

The two alternative allocations further support this conclusion. The accuracy-oriented variant, which substantially increases the weight on $R^\text{acc}$, leads to clear drops across nearly all metrics, especially in visual grounding. Likewise, the format-enhanced allocation $(0.20, 0.20, 0.30, 0.30, 0.30)$, which assigns a larger coefficient to the structural reward than the adopted configuration, also remains weaker overall. These results indicate that the FTPO objective benefits from keeping the format reward relatively small while explicitly prioritizing $R^\text{box}$ and $R^\text{txt}$. Overall, the chosen reward allocation is not arbitrary, but a task-aligned design that better matches the core requirements of multimodal sarcasm target identification.

\begin{figure*}[t]
\centering
\begin{minipage}{0.95\textwidth}
\begin{promptpanel}{Prompt for Data Distillation}
[
# Role
  You are a **Ground Truth Explainer**. Your task is to generate a logical reasoning chain that justifies the provided Ground Truth Label.
# INPUTS
  1. Tweet Text
  2. **FIXED Label**: (0 or 1)
  3. **Target Box**: `[x1, y1, x2, y2]`.
    - Note: If the box is `[0, 0, 1000, 1000]`, it means the **WHOLE IMAGE** is relevant.
    - If the box is smaller, it focuses on a specific object.
# INSTRUCTION
  *   **DO NOT PREDICT** the label. Just explain WHY the provided label is correct.
  *   **FORCE ALIGNMENT**:
      - If Label 1 (Sarcastic): Explain the contradiction between the Text and the Image contents (inside the Target Box).
      - If Label 0 (Not Sarcastic): Explain the alignment/neutrality.
# Output Format (Use <Think> and <Answer>)
<Think>
(Reasoning)
  1. Visual Description:
    - If box is [0,0,1000,1000]: Describe the **overall scene** and main atmosphere.
    - If box is specific: Describe strictly the **object inside the box**.
  2. Text Analysis: Literal meaning of the tweet.
  3. Incongruity Check: Explain the relationship (Contradiction or Agreement).
</Think>
<Answer>
  (Placeholder - will be overwritten by code)
  Label: "..."
  Image Objects: [...]
  Text Objects: "..."
</Answer>
]
\end{promptpanel}
\end{minipage}
\Description{Full-width appendix prompt panel for data distillation with a pale blue background and centered title tab.}
\caption{Prompt used to distill Grounded CoT reasoning annotations from the teacher model.}
\label{fig:distill_prompt}
\end{figure*}

\clearpage

\begin{figure*}[t]
\centering
\begin{minipage}{0.95\textwidth}
\begin{promptpanel}{Prompt for Model Inference}
[
# Role
  You are an expert in **Multimodal Sarcasm Detection**. Your goal is to determine if a tweet (Text + Image) is sarcastic by identifying contradictions between visual and textual information.
# Task
  Analyze the input image and text step-by-step. You must output your reasoning process in a <Think> block, followed by a structured final answer in an <Answer> block.
# Response Format
  ## 1. Visual-Textual Reasoning (<Think>)
    Inside <Think>, strictly follow these steps:
      1.  **Visual grounding**:
          *   List key objects in the image.
          *   **CRITICAL**: You MUST append normalized coordinates `[xmin, ymin, xmax, ymax]` (0-1000) immediately after mentioning a relevant object.
      2.  **Semantic Analysis**:
          *   Analyze the literal meaning and emotional tone of the tweet text.
      3.  **Incongruity Check**:
          *   Compare the Visual Reality with the Text Description.
          *   **Explicitly state**: "Does the image support the text, or contradict it?"
  ## 2. Final Structured Output (<Answer>)
    Inside `<Answer>`, output ONLY the following valid JSON-like format:
      Label: "sarcastic" OR "not sarcastic"
      Image Objects: [(xmin,ymin,xmax,ymax)]
      Text Objects: "keyword1, keyword2"
# Constraints & Rules
  1.  **Label**: strictly "sarcastic" or "not sarcastic".
  2.  **Image Objects (Crucial Rules)**:
    *   If a **specific object** contradicts the text, output its bounding box (e.g., `[(200,300,500,600)]`).
    *   If the **Whole Scene** or **General Atmosphere** contradicts the text (e.g., bad weather, traffic jam, crowd), use **Full Image Coordinates**: `[(0,0,1000,1000)]`.
    *   If not sarcastic, output `[(0,0,0,0)]` or empty brackets `[]`.
  3.  **Text Objects**:
    *   Extract only the 1-3 keywords triggering the irony.
    *   If not sarcastic, output `""`.
]
\end{promptpanel}
\end{minipage}
\Description{Full-width appendix prompt panel for model inference with a pale blue background and centered title tab.}
\caption{Prompt used during inference for grounded multimodal sarcasm reasoning and structured prediction.}
\label{fig:infer_prompt}
\end{figure*}

\clearpage

\begin{figure*}[t]
\centering
\begin{minipage}{0.95\textwidth}
\begin{promptpanel}{Prompt for LLM-as-a-Judge Evaluation}
[
  Task: Evaluate a model's reasoning for Multimodal Sarcasm Detection. Be a STRICT judge. 5 is extremely rare.
  [Inputs]
    1. Image
    2. Text: {text_input}
    3. Ground Truth: Label=[{gt_label}], Box(0-1000)=[{gt_boxes}], Words=[{gt_words}]
    4. Model Output: {model_response}
  [Scoring 1-5]
    - V_Score (Visual): 1=hallucinated/missed GT box, 3=superficial, 5=perfectly identified GT objects.
    - R_Score (Reasoning): 1=wrong logic, 3=shallow textual analysis, 5=deep text-image contradiction analysis matching GT.
    - C_Score (Consistency): 1=conclusion contradicts reasoning, 5=perfectly aligned.
  Output pure JSON only, no markdown, no explanations:
    {
      "V": <int>, "R": <int>, "C": <int>
    }
]
\end{promptpanel}
\end{minipage}
\Description{Full-width appendix prompt panel for LLM-as-a-Judge evaluation with a pale blue background and centered title tab.}
\caption{Prompt used by the LLM-as-a-Judge evaluator for assessing reasoning quality.}
\label{fig:judge_prompt}
\end{figure*}

\end{document}